\pdfoutput=1
\documentclass{article} 
\usepackage{iclr-style/iclr2016_conference,times}
\usepackage{hyperref}
\usepackage{url}
\usepackage{amsmath}
\usepackage{amsfonts}
\usepackage{graphicx}
\usepackage{caption}
\usepackage{subfig}
\usepackage{tabulary}
\usepackage{soul}
\usepackage{color}
\usepackage{xcolor}
\usepackage[bottom]{footmisc}

\newcommand{\Figref}[1]{Fig.~\ref{#1}}

\newcommand{\Eqref}[1]{Eq.~\ref{#1}} 

\def\beqa#1\eeqa{\begin{eqnarray}#1\end{eqnarray}}

\newcommand{\comm}[1]{}
\newcommand{\given}{\,|\,}
\newcommand{\expectation}{\mathbb{E}}
\newcommand{\kldiv}{\mathrm{D}_{\rm KL}}
\newcommand{\klBars}{\,\|\,}
\newcommand{\sigmoid}{\boldsymbol{\sigma}}
\newcommand{\hlang}{h^{lang}}
\newcommand{\hlangall}{h^{lang}}
\newcommand{\hdec}{h^{gen}}

\newcommand{\henc}{h^{infer}}
\newcommand{\readop}{\mathit{read}}
\newcommand{\writeop}{\mathit{write}}
\newcommand{\encoder}{\mathit{LSTM}^{infer}}
\newcommand{\decoder}{\mathit{LSTM}^{gen}}
\newcommand{\canv}{c}
\newcommand{\lat}{z}
\newcommand{\vv}{v}
\newcommand{\Lat}{Z}
\newcommand{\numSamples}{L}
\newcommand{\sampleIdx}{l}
\newcommand{\LatSample}{\tilde{z}}
\newcommand{\icaption}{{\bf{y}}}
\newcommand{\oimage}{{\bf{x}}}
\newcommand{\post}{Q}
\newcommand{\prior}{P}
\newcommand{\loss}{\mathcal{L}}

\newcommand{\hlc}[2][yellow]{{\sethlcolor{#1}\hl{#2}}}
\newcommand{\gFilterx}{F_x}
\newcommand{\gFiltery}{F_y}
\newcommand{\WriteFunc}{K}
\newcommand{\real}{\mathbb{R}}

\definecolor{0}{RGB}{247,251,255}
\definecolor{1}{RGB}{222,235,247}
\definecolor{2}{RGB}{198,219,239}
\definecolor{3}{RGB}{158,202,225}
\definecolor{4}{RGB}{107,174,214}
\definecolor{5}{RGB}{66,146,198}

\newcommand{\hlczero}[1]{\hlc[0]{#1}}
\newcommand{\hlcone}[1]{\hlc[1]{#1}}
\newcommand{\hlctwo}[1]{\hlc[2]{#1}}
\newcommand{\hlcthree}[1]{\hlc[3]{#1}}

\newenvironment{myfont}{\fontfamily{[pcr]}\selectfont}{\par}
\DeclareTextFontCommand{\textmyfont}{\myfont}

\iclrfinalcopy

\title{Generating Images from Captions\\ with Attention}

\author{
Elman Mansimov, Emilio Parisotto, Jimmy Lei Ba \& Ruslan Salakhutdinov\\
Department of Computer Science\\
University of Toronto\\
Toronto, Ontario, Canada\\
\texttt{\{emansim,eparisotto,rsalakhu\}@cs.toronto.edu}, \texttt{jimmy@psi.utoronto.ca}
}

\begin{document}

\maketitle

\begin{abstract}
Motivated by the recent progress in generative models, we introduce a model that generates images from natural language descriptions. The proposed model iteratively draws patches on a canvas, while attending to the relevant words in the description. After training on Microsoft COCO, we compare our model with several baseline generative models on image generation and retrieval tasks. We demonstrate that our model produces higher quality samples than other approaches and generates images with novel scene compositions corresponding to previously unseen captions in the dataset. 
\end{abstract}

\section{Introduction}
\vspace{-0.1in}
Statistical natural image modelling remains a fundamental problem in computer vision and image understanding. 
The challenging nature of this task has motivated recent approaches to exploit the inference and generative capabilities of deep neural networks.
Previously studied deep generative models of images often defined distributions that were restricted to being either unconditioned or conditioned on classification labels. In real world applications, however, images rarely appear in isolation as they are often accompanied by unstructured textual descriptions, such as on web pages and in books. 
The additional information from these descriptions could be used to simplify the image modelling task. Moreover, learning generative models conditioned on text also allows a better understanding of the generalization performance of the model, as we can create textual descriptions of completely new scenes not seen at training time. 

There are numerous ways to learn a generative model over both image and text modalities. One approach is to learn a generative model of text conditioned on images, known as caption generation
\citep{kiros_icml14,karpathy_captions,vinyals_captions,xu_captions}. These models take an image descriptor and generate unstructured texts using a recurrent decoder. In contrast, in this paper we explore models that condition in the opposite direction, i.e. taking textual descriptions as input and using them to generate relevant images.
Generating high dimensional realistic images from their descriptions 
combines the two challenging components of language modelling and image generation, and can be considered to be more difficult than caption generation. 

\begin{figure}[!t]
\captionsetup[subfigure]{labelformat=empty}
\vspace{-0.3in}
\begin{center}
\subfloat[A stop sign is flying in blue skies.
]{\includegraphics[width=0.23\textwidth]{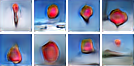}}\quad
\subfloat[A herd of elephants flying in the blue skies.
]{\includegraphics[width=0.23\textwidth]{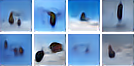}}\quad
\subfloat[A toilet seat sits open in the grass field.
]{\includegraphics[width=0.23\textwidth]{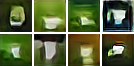}}\quad
\subfloat[A person skiing on sand clad vast desert.
]{\includegraphics[width=0.23\textwidth]{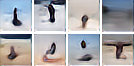}}\quad
\end{center}
\caption{\small 
Examples of generated images based on captions that describe 
novel scene compositions that are highly unlikely to occur in real life. 
The captions describe a common object doing unusual things or set in a strange location. 
}
\label{fig:genimages4}
\vspace{-0.2in}
\end{figure}

In this paper, we illustrate how sequential deep learning techniques can be used to build a conditional probabilistic model over natural image space effectively. By extending the Deep Recurrent Attention Writer (DRAW) \citep{gregor_draw}, our model iteratively draws patches on a canvas, while attending to the relevant words in the description. Overall, the main contributions of this work are the following: we introduce a conditional alignDRAW model, a generative model of images from captions using a soft attention mechanism. The images generated by our alignDRAW model are refined in a post-processing step by a deterministic Laplacian pyramid adversarial network \citep{denton_lapgan}. We further 
illustrate how our method, learned on Microsoft COCO~\citep{mscoco}, generalizes to captions describing novel scenes that are not 
seen in the dataset, such as ``A stop sign is flying in blue skies'' (see \Figref{fig:genimages4}). 

\section{Related Work}
\vspace{-0.1in}
Deep Neural Networks have achieved significant success in various tasks such as image recognition \citep{krizhevsky_imagenet}, speech transcription \citep{graves_speech}, and machine 
translation~\citep{bahdanau_mt}. 
While most of the recent success has been achieved by discriminative models, generative models have not yet enjoyed the same level of success. Most of the previous work in generative models has focused on variants of Boltzmann Machines \citep{smolensky_rbm,russ_dbm} and Deep Belief Networks~\citep{hinton_dbn}. While these models are very powerful, each iteration of training requires a computationally costly step of MCMC to approximate derivatives of an 
intractable partition function (normalization constant), making it difficult to scale them to large datasets.

\cite{kingma_vae}, \cite{rezende_vae} have introduced the Variational Auto-Encoder (VAE) which can be seen as a neural network with continuous latent variables. The encoder is used to approximate a posterior distribution and the decoder is used to stochastically reconstruct the data from latent variables. 
\cite{gregor_draw} further introduced the Deep Recurrent Attention Writer (DRAW), extending the VAE approach by incorporating a novel differentiable attention mechanism.

Generative Adversarial Networks (GANs) \citep{goodfellow_gan} are another type of generative models that use noise-contrastive estimation \citep{gutmann_nce} to avoid calculating an intractable partition function. The model consists of a generator that generates samples using a uniform distribution and a discriminator that discriminates between real and generated images. 
Recently, \cite{denton_lapgan} have scaled those models by training conditional GANs at each level of a Laplacian pyramid of images. 

While many of the previous approaches have focused on unconditional models or models conditioned on labels, 
in this paper we develop a generative model of images conditioned on captions.

\section{Model}
\vspace{-0.1in}
\label{sec:model}
Our proposed model 
defines a generative process of images conditioned on captions. In particular, captions are represented as a sequence of consecutive words and images are represented as a sequence of patches drawn on a canvas $c_t$ over time $t=1,...,T$. The model can be viewed as a part of the sequence-to-sequence framework \citep{ilya_mt,cho_mt,nitish_video}.

\subsection{Language Model: the Bidirectional Attention RNN}
\vspace{-0.05in}
\label{sec:lang}
Let $\icaption$ be the input caption, represented as a sequence
of 1-of-K encoded words 
${\bf y} = (y_{1}, y_{2}, ..., y_{N})$, 
where $K$ is the size of the vocabulary and $N$ is the length of the sequence.
We obtain the caption sentence representation by first 
transforming each word $y_{i}$ to an $m$-dimensional 
vector representation $\hlang_{i}$, $i=1,..,N$ using the Bidirectional RNN. In a Bidirectional RNN, the two LSTMs~\citep{hochreiter_lstm} with \textit{forget} gates 
\citep{gers_forget} process the input sequence from both forward and backward directions. The Forward LSTM computes the sequence of forward hidden states $[\overrightarrow{h}^{lang}_{1}, \overrightarrow{h}^{lang}_{2}, ..., \overrightarrow{h}^{lang}_{N}]$ , whereas the Backward LSTM computes the sequence of backward hidden states $[\overleftarrow{h}^{lang}_{1}, \overleftarrow{h}^{lang}_{2}, ..., \overleftarrow{h}^{lang}_{N}]$. 
These hidden states are then concatenated together 
into the sequence $\hlangall = [\hlang_{1}, \hlang_{2}, ..., \hlang_{N}]$, 
with $\hlang_{i} = [\overrightarrow{h}^{lang}_{i}, \overleftarrow{h}^{lang}_{i}], 1\leq i\leq N$.

\begin{figure}[t!]
\vspace{-0.3in}
\captionsetup[subfigure]{labelformat=empty}
\begin{center}
\includegraphics[width=0.99\textwidth]{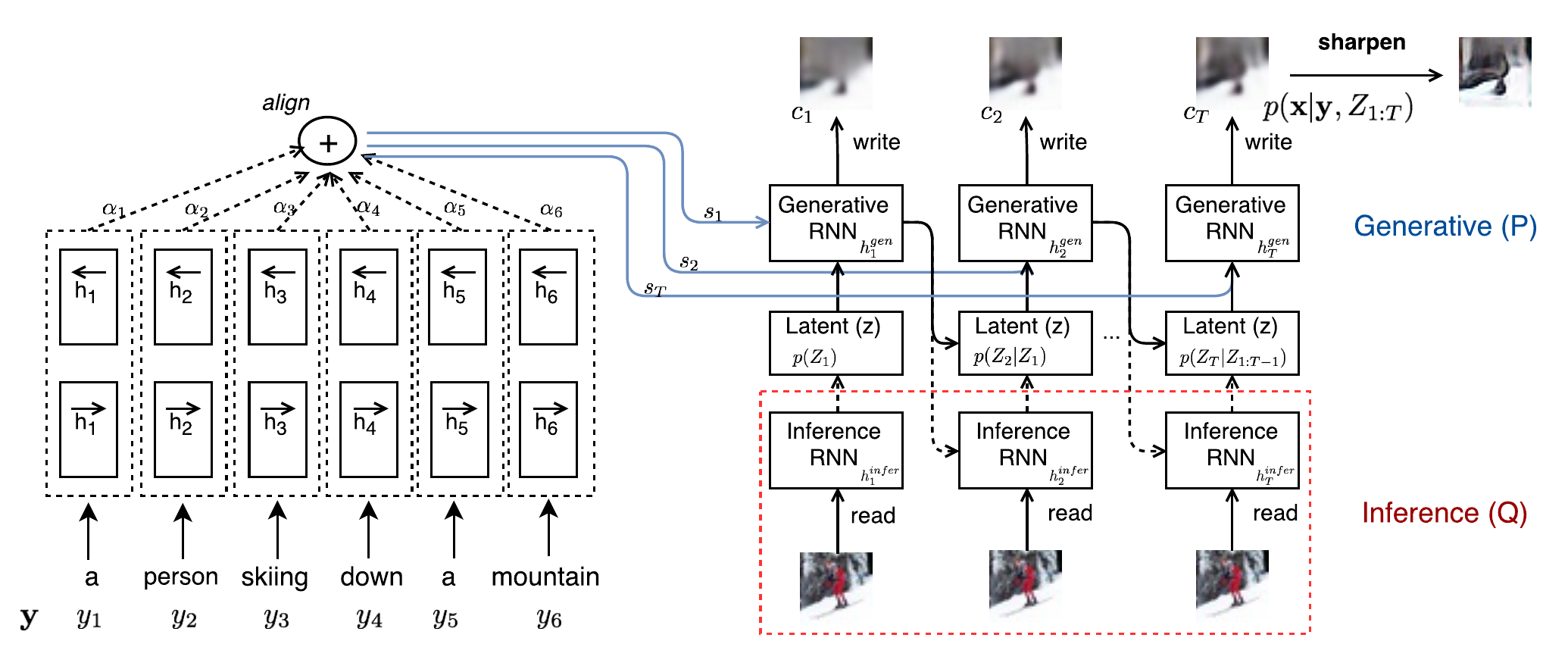}\quad
\end{center}
\vspace{-0.1in}
\caption{\small AlignDRAW model for generating images by learning an alignment between the input captions
and generating canvas. The caption is encoded using 
the Bidirectional RNN (left). The generative RNN takes a latent sequence $z_{1:T}$ sampled from the 
prior along with the dynamic caption representation $s_{1:T}$ to generate the canvas matrix $c_T$, which is then
used to generate the final image $\oimage$ (right).
The inference RNN is used to compute approximate posterior $Q$ over the latent sequence.} 
\label{fig:figmodel}
\vspace{-0.2in}
\end{figure}

\subsection{Image Model: the Conditional DRAW Network}
\vspace{-0.05in}
\label{sec:image_model}
To generate an image $\oimage$ conditioned on the caption information $\bf{y}$,
we extended the DRAW network~\citep{gregor_draw} to include caption
representation $\hlangall$ at each step, as shown in \Figref{fig:figmodel}.
The conditional DRAW network is a stochastic recurrent neural network
that consists of a sequence of latent variables $\Lat_t \in \mathbb{R}^D$,
$t=1,..,T$, where the output is accumulated over all $T$ time-steps. For
simplicity in notation, the images $\oimage\in\real^{h\times w}$ are assumed to have size $h$-by-$w$ and only one color channel.

Unlike the original DRAW network where latent variables are independent 
spherical Gaussians $\mathcal{N}(0, I)$, the latent variables in the proposed alignDRAW model have their mean and variance depend on the previous hidden states of the generative LSTM $\hdec_{t-1}$, except for $\prior(\Lat_1) = \mathcal{N}(0, I)$.  
Namely, the mean and variance of the prior distribution over $\Lat_t$ are parameterized by: 
\begin{align}
\prior(\Lat_t|\Lat_{1:t-1}) &= \mathcal{N}\bigg(\mu(\hdec_{t-1}), \sigma(\hdec_{t-1})\bigg), \nonumber \\
\mu(\hdec_{t-1}) &= \tanh(W_{\mu}\hdec_{t-1}),\nonumber \\
\sigma(\hdec_{t-1}) &= \exp\big(\tanh(W_{\sigma}\hdec_{t-1})\big), \nonumber 
\end{align}
where $W_{\mu} \in \mathbb{R}^{D \times n}$, $W_{\sigma} \in \mathbb{R}^{D \times n}$ are 
the learned model parameters, and $n$ is the dimensionality of $\hdec_{t}$,
the hidden state of the generative LSTM. 
Similar to \citep{bachman_sdm}, we have observed that 
the model performance is improved by including dependencies between latent variables. 

Formally, an image is generated by iteratively computing the following set of 
equations for $t=1,...,T$ (see \Figref{fig:figmodel}), with $\hdec_0$ and $c_0$
initialized to learned biases:
\begin{align}
\label{eq:x_hat}
\lat_t &\sim \prior(\Lat_t|\Lat_{1:t-1}) = \mathcal{N}\bigg(\mu(\hdec_{t-1}), \sigma(\hdec_{t-1})\bigg),\\
s_{t} &= align(\hdec_{t-1}, \hlangall), \\
\label{eq:decoder}
\hdec_t &= \decoder(\hdec_{t-1}, [z_t, s_{t}]), \\
\label{eq:writeop}
\canv_t &= \canv_{t-1} + \writeop(\hdec_t), \\
\tilde{\oimage} &\sim P(\oimage\given\icaption, \Lat_{1:T}) = \prod_i P(x_i\given\icaption, \Lat_{1:T}) = \prod_i \textrm{Bern}(\sigmoid(\canv_{T,i})). 
\label{eq:write}
\end{align}
The $align$ function is used to compute the alignment between the input caption and intermediate image generative steps \citep{bahdanau_mt}. 
Given the caption representation from the language model, $\hlangall = [\hlang_{1}, \hlang_{2}, ..., \hlang_{N}]$, the $align$ operator outputs a dynamic sentence representation $s_t$ at each step through a weighted sum using alignment probabilities 
$\alpha_{1...N}^{t}$:
\beqa
s_t=align(\hdec_{t-1}, \hlangall) = 
\alpha_{1}^{t}\hlang_{1} + \alpha_{2}^{t}\hlang_{2} + ... + \alpha_{N}^{t}\hlang_{N}.
\eeqa
The corresponding alignment probability $\alpha_{k}^{t}$ for the $k^{th}$ word in the caption is obtained using the caption representation $\hlangall$ and the current hidden state of the generative model $\hdec_{t-1}$: 
\beqa
\label{eq:align2}
\alpha_{k}^{t} = \frac{\exp\left(\vv^{\top}\tanh(U\hlang_{k} + W\hdec_{t-1} + b)\right)}
  {\sum_{i=1}^N \exp\left(\vv^{\top}\tanh(U\hlang_{i} + W\hdec_{t-1} + b)\right)},
\eeqa
where $\vv \in \mathbb{R}^{l}$, $U \in \mathbb{R}^{l \times m}$, $W \in \mathbb{R}^{l \times n}$ and 
$b \in \mathbb{R}^{l} $ are the learned model parameters of the alignment model. 

The $\decoder$ function of \Eqref{eq:decoder} 
is defined by the LSTM network with \textit{forget} gates 
\citep{gers_forget} at a single time-step. To generate the next 
hidden state $\hdec_t$, 
the $\decoder$ takes the previous hidden state $\hdec_{t-1}$ and
combines it with the input from both the latent sample $z_t$ and the sentence representation $s_t$.

The output of the $\decoder$ function $\hdec_t$ is then passed through the $\writeop$ operator 
which is added to a cumulative canvas matrix $c_t \in \real^{h \times w}$ (\Eqref{eq:writeop}). 
The $\writeop$ operator produces two arrays of 1D Gaussian filter banks $\gFilterx(\hdec_t)\in\real^{h\times p}$ and $\gFiltery(\hdec_t)\in\real^{w\times p}$ whose filter locations and scales are computed from the generative LSTM hidden state $\hdec_t$ (same as defined in \cite{gregor_draw}). 
The Gaussian filter banks are then applied to the generated $p$-by-$p$ image patch $\WriteFunc(\hdec_t) \in \real^{p\times p}$, placing it onto the canvas:
\beqa
\label{eq:write}
\Delta \canv_{t} = \canv_{t} - \canv_{t-1} = \writeop(\hdec_t) = \gFilterx(\hdec_t) \WriteFunc(\hdec_t) \gFiltery(\hdec_t)^\top. 
\eeqa

Finally, each entry $\canv_{T,i}$ from the final canvas matrix $\canv_T$ is transformed using a sigmoid function $\sigmoid$ to produce a conditional Bernoulli 
distribution with mean vector $\sigmoid(\canv_{T})$ over the 
$h\times w$ image pixels~$\oimage$ given the latent variables 
$\Lat_{1:T}$ and the input caption $\icaption$\footnote{We also experimented with a conditional Gaussian observation model, but
it worked worse compared to the Bernoulli model.}. In practice, when generating an 
image $\oimage$, instead of sampling from the conditional Bernoulli
distribution, we simply use the conditional mean $\oimage = \sigmoid(\canv_{T})$.

\subsection{Learning}
\vspace{-0.05in}
\label{sec:learning}
The model is trained to maximize a variational lower bound $\loss$ 
on the marginal likelihood of the correct image $\oimage$ given the input caption $\icaption$:
\beqa
\label{eq:varbound}
\loss = \sum_{\Lat}Q(\Lat\given\oimage,\icaption) \log P(\oimage\given\icaption, \Lat) - \kldiv\left(Q(\Lat\given\oimage,\icaption)\klBars 
  P(\Lat \given\icaption)\right) \le \log P(\oimage\given\icaption).
\eeqa
Similar to the DRAW model, the inference recurrent network 
produces an approximate posterior $Q(\Lat_{1:T}\given\oimage,\icaption)$ via a $\readop$ operator, which reads a patch from an input image $\oimage$ using two arrays of 1D Gaussian filters (inverse of $\writeop$ from section \ref{sec:image_model}) at each time-step $t$. Specifically, 
\begin{align}
\label{eq:x_hat}
\hat{\oimage}_t &= \oimage-\sigmoid(\canv_{t-1}),\\
\label{eq:read}
r_t &= \readop(\oimage_t, \hat{\oimage}_t, \hdec_{t-1}),\\
\henc_t &= \encoder(\henc_{t-1}, [r_t, \hdec_{t-1}]),\\
\post(\Lat_t|\oimage,\icaption,\Lat_{1:t-1}) &= \mathcal{N}\left(\mu(\henc_t), \sigma(\henc_t)\right),
\end{align}
where $\hat{\oimage}$ is the \textit{error} image and $\henc_0$ is initialized to the learned bias $b$. 
Note that the inference $\encoder$ 
takes as its input both the output
of the $\readop$ operator $r_t\in \real^{p\times p}$, which depends on the original input image $\oimage$,
and the previous state of the 
generative decoder $\hdec_{t-1}$, which depends on the latent sample history $z_{1:t-1}$ and 
dynamic sentence representation $s_{t-1}$ (see \Eqref{eq:decoder}). 
Hence, the approximate posterior $Q$ will depend on the input image $\oimage$,
the corresponding caption $\icaption$, and the latent history $\Lat_{1:t-1}$, except  
for the first step $Q(Z_1 | \oimage)$, which depends only on $\oimage$.  

The terms in the variational lower bound Eq. \ref{eq:varbound} can be rearranged using the law of total expectation. Therefore, the variational bound $\loss$ is calculated as follows:
\begin{align}
\label{eq:loss}
\loss =  &\expectation_{Q(\Lat_{1:T}\given\icaption,\oimage)}\left[ \log p(\oimage\given\icaption,\Lat_{1:T}) - \sum_{t=2}^T\kldiv\left(\post(\Lat_t\given\Lat_{1:t-1},\icaption,\oimage)\klBars\prior(\Lat_t\given\Lat_{1:t-1},\icaption)\right) \right] \nonumber \\& - \kldiv\left(\post(\Lat_1\given\oimage)\klBars\prior(\Lat_1)\right).
\end{align}
The expectation can be approximated by $\numSamples$ Monte Carlo samples $\LatSample_{1:T}$ from $\post(\Lat_{1:T}\given\icaption, \oimage)$:
\begin{align}
\label{eq:loss_mc}
\loss \approx  &\frac{1}{\numSamples}\sum_{\sampleIdx=1}^\numSamples\left[ \log p(\oimage\given\icaption,\LatSample^\sampleIdx_{1:T}) - \sum_{t=2}^T\kldiv\left(\post(\Lat_t\given\LatSample^\sampleIdx_{1:t-1},\icaption,\oimage)\klBars\prior(\Lat_t\given\LatSample^\sampleIdx_{1:t-1},\icaption)\right) \right] \nonumber \\& - \kldiv\left(\post(\Lat_1\given\oimage)\klBars\prior(\Lat_1)\right).
\end{align}
The model can be trained using stochastic gradient descent. 
In all of our experiments, we used only a single sample 
from $\post(\Lat_{1:T}\given\icaption, \oimage)$ for parameter learning.
Training details, hyperparameter settings,
and the overall model architecture are specified in Appendix B. 
The code is available at \url{https://github.com/emansim/text2image}.

\subsection{Generating Images from Captions}
\vspace{-0.05in}
During the image generation step, 
we discard the inference network and instead sample from the prior distribution. 
Due to the blurriness of samples generated by the DRAW model, we perform an additional post processing step where we use an 
adversarial network trained on residuals of a Laplacian pyramid conditioned on the skipthought representation \citep{kiros_skipthought} of the captions 
to sharpen the generated images, similar to \citep{denton_lapgan}. By fixing the prior of the adversarial generator to its mean, it gets treated as a deterministic neural network that allows us to define the conditional data term in \Eqref{eq:loss} on the sharpened images and 
estimate the variational lower bound accordingly. 

\begin{figure}[!t]
\captionsetup[subfigure]{labelformat=empty}
\vspace{-0.3in}
\begin{center}
\subfloat[A \underline{yellow} school bus parked in a parking lot.
]{\includegraphics[width=0.23\textwidth]{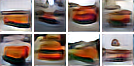}}\quad
\subfloat[A \underline{red} school bus parked in a parking lot.
]{\includegraphics[width=0.23\textwidth]{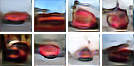}}\quad
\subfloat[A \underline{green} school bus parked in a parking lot.
]{\includegraphics[width=0.23\textwidth]{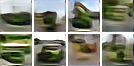}}\quad
\subfloat[A \underline{blue} school bus parked in a parking lot.
]{\includegraphics[width=0.23\textwidth]{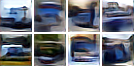}}\quad
\\
\subfloat[\underline{The decadent chocolate} \underline{desert} is on the table.
]{\includegraphics[width=0.23\textwidth]{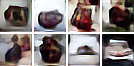}}\quad
\subfloat[\underline{A bowl of bananas} is on the table.
]{\includegraphics[width=0.23\textwidth]{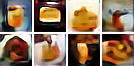}}\quad
\subfloat[A vintage photo of a \underline{cat}.
]{\includegraphics[width=0.23\textwidth]{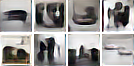}}\quad
\subfloat[A vintage photo of a \underline{dog}.
]{\includegraphics[width=0.23\textwidth]{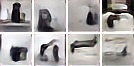}}\quad
\end{center}
\caption{\small
 \textbf{Top}: Examples of changing the color while keeping the caption fixed.
 \textbf{Bottom}: Examples of changing the object while keeping the caption fixed.
 The shown images are the probabilities $\sigmoid(\canv_{T})$. Best viewed in colour.}
\label{fig:genimages3}
\vspace{-0.2in}
\end{figure}

\section{Experiments}
\vspace{-0.05in}
\subsection{Microsoft COCO}
\vspace{-0.05in}
Microsoft COCO \citep{mscoco} is a large dataset containing 82,783 images, each annotated with at least 5 captions. The rich collection of images with a wide variety of styles, backgrounds and objects makes the task of learning a good generative model 
very challenging. For consistency with related work on caption generation, we used only the first five captions when training and evaluating our model. 
The images were resized to $32 \times 32$ pixels for consistency with other tiny image datasets \citep{krizhevsky_cifar}. In the following subsections, we analyzed both the qualitative and quantitative aspects of our model as well as compared its performance with that of other, related generative models.\footnote{To see more
generated images, visit \url{http://www.cs.toronto.edu/~emansim/cap2im.html}}
Appendix A further reports some additional experiments using the MNIST dataset. 

\subsubsection{Analysis of Generated Images}
\vspace{-0.05in}
The main goal of this work is to learn a model that can understand the semantic meaning expressed in the textual descriptions of images, such as the properties of objects, the relationships between them, and then use that knowledge to generate relevant images. To examine the understanding of our model, we wrote a set of captions inspired by the COCO dataset and changed some words in the captions to see whether the model made the relevant changes in the generated samples.

First, we explored whether the model understood one of the most basic properties of any object, the color. In \Figref{fig:genimages3}, we generated images of school buses with four different colors: yellow, red, green and blue. Although, there are images of buses with different colors in the training set, all mentioned school buses are specifically colored yellow. Despite that, the model managed to generate images of an object that is visually reminiscent of a school bus that is painted with the specified color.

\begin{figure}[!t]
\captionsetup[subfigure]{labelformat=empty}
\vspace{-0.2in}
\begin{center}
\subfloat[]{\includegraphics[width=0.23\textwidth]{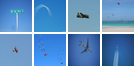}}\quad
\subfloat[]{\includegraphics[width=0.23\textwidth]{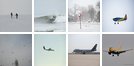}}\quad
\subfloat[]{\includegraphics[width=0.23\textwidth]{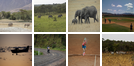}}\quad
\subfloat[]{\includegraphics[width=0.23\textwidth]{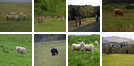}}\\
\vspace{-0.45cm}
\subfloat[A very large commercial plane flying in \underline{blue} skies.
]{\includegraphics[width=0.23\textwidth]{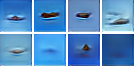}}\quad
\subfloat[A very large commercial plane flying in \underline{rainy} skies.
]{\includegraphics[width=0.23\textwidth]{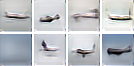}}\quad
\subfloat[A herd of elephants walking across a \underline{dry} grass field.
]{\includegraphics[width=0.23\textwidth]{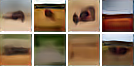}}\quad
\subfloat[A herd of elephants walking across a \underline{green} grass field.
]{\includegraphics[width=0.23\textwidth]{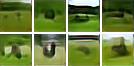}}\quad
\end{center}
\caption{\textbf{Bottom}: Examples of changing the background while keeping the caption fixed. \textbf{Top}: The respective nearest training images based on pixel-wise L2 distance. The nearest images from the training set also indicate that the model was not simply copying the patterns it observed during the learning phase.}
\label{fig:genimages2}
\vspace{-0.1in}
\end{figure}

Apart from changing the colors of objects, we next examined whether changing the background of the scene described in a caption would result in the appropriate changes in the generated samples. The task of changing the background of an image is somewhat harder than just changing the color of an object because the model will have to make alterations over a wider visual area. Nevertheless, as shown in \Figref{fig:genimages2} changing the skies from blue to rainy in a caption as well as changing the grass type from dry to green in another caption resulted in the appropriate changes in the generated image.

Despite a large number of ways of changing colors and backgrounds in descriptions, in general we found that the model made appropriate changes as long as some similar pattern was present in the training set. However, the model struggled when the visual difference between objects was very small, such as when the objects have the same general shape and color. In \Figref{fig:genimages3}, we demonstrate that when we swap two objects that are both visually similar, for example cats and dogs, it is difficult to discriminate solely from the generated samples whether it is an image of a cat or dog, even though we might notice an animal-like shape. This highlights a limitation of the model in that it has difficulty modelling the fine-grained details of objects.

As a test of model generalization, we tried generating images corresponding to captions that describe scenarios that are highly unlikely to occur in real life. These captions describe a common object doing unusual things 
or set in a strange location, for example ``A toilet seat sits open in the grass field''.
Even though some of these scenarios may never occur in real life, it is very easy for humans to imagine the corresponding scene. Nevertheless, as you can see in \Figref{fig:genimages4}, the model managed to generate reasonable images.  

\subsubsection{Analysis of Attention}
\begin{figure}[!t]
\captionsetup[subfigure]{labelformat=empty}
\vspace{-0.3in}
\begin{center}
\subfloat[\hlcthree{A} rider \hlcone{on} a blue \hlcone{motorcycle} in the \underline{\hlctwo{desert}}.
]{\includegraphics[width=0.23\textwidth]{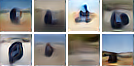}}\quad
\subfloat[\hlcthree{A} rider \hlcone{on} a blue \hlcone{motorcycle} in the \underline{\hlctwo{forest}}.
]{\includegraphics[width=0.23\textwidth]{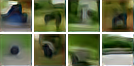}}\quad
\subfloat[\hlctwo{A} \hlcone{surfer}, a woman, and a child walk on the \underline{\hlctwo{beach}}.
]{\includegraphics[width=0.23\textwidth]{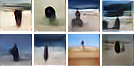}}\quad
\subfloat[\hlcthree{A} \hlcone{surfer}, a woman, and a child walk on the \underline{\hlczero{sun}}.
]{\includegraphics[width=0.23\textwidth]{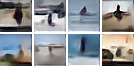}}\quad
\\
\subfloat[alignDRAW]{\includegraphics[width=0.23\textwidth]{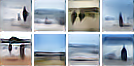}}\quad
\subfloat[LAPGAN]{\includegraphics[width=0.23\textwidth]{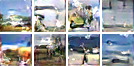}}\quad
\subfloat[Conv-Deconv VAE]{\includegraphics[width=0.23\textwidth]{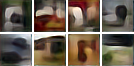}}\quad
\subfloat[Fully-Conn VAE]{\includegraphics[width=0.23\textwidth]{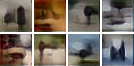}}\quad
\end{center}
\caption{ \textbf{Top}: Examples of most attended words while changing the background in the caption. 
\textbf{Bottom}: Four different models displaying results from sampling caption \textit{A group of people walk on a beach with surf boards.}}
\label{fig:diffmodels}
\vspace{-0.2in}
\end{figure}

\vspace{-0.05in}
After flipping sets of words in the captions, we further explored which words the model attended to when generating images. It turned out that during the generation step, the model mostly focused on the specific words (or nearby words) that carried the main semantic meaning expressed in the sentences. The attention values of words in sentences helped us interpret the reasons why the model made the changes it did when we flipped certain words. For example, in \Figref{fig:diffmodels}, top row, we can see that when we flipped the word ``desert'' to ``forest'', the attention over words in the sentence did not change drastically. This suggests that, in their respective sentences, the model looked at ``desert'' and ``forest'' with relatively equal probability, and thus made the correct changes. In contrast, when we swap words ``beach'' and ``sun'', we can see a drastic change between sentences in the probability distribution over words. By noting that the model completely ignores the word ``sun'' in the second sentence, we can therefore gain a more thorough understanding of why we see no visual differences between the images generated by each caption.

We also tried to analyze the way the model generated images. Unfortunately, we found that there was no significant connection between the patches drawn on canvas and the most attended words at particular time-steps.

\subsubsection{Comparison With Other Models}
\vspace{-0.05in}
Quantitatively evaluating generative models remains a challenging task in itself as each method of evaluation suffers from its own specific drawbacks. Compared to reporting classification accuracies in discriminative models, the measures defining generative models are intractable most of the times and might not correctly define the real quality of the model. To get a better comparison between performances of different generative models, we report results on two different metrics as well as a qualitative comparison of different generative models.

We compared the performance of the proposed model to the DRAW model conditioned on captions without the $align$ function (noalignDRAW) as well as the DRAW model conditioned on the skipthought vectors of \citep{kiros_skipthought} (skipthoughtDRAW). All of the conditional DRAW models were trained with a binary cross-entropy cost function, i.e. they had Bernoulli conditional likelihoods. We also compared our model with Fully-Connected (Fully-Conn) and Convolutional-Deconvolutional (Conv-Deconv) Variational Autoencoders which were trained with the least squares cost function. The LAPGAN model of~\citep{denton_lapgan} was trained on a two level Laplacian Pyramid with a GAN as a top layer generator and all stages were conditioned on the same skipthought vector.

In \Figref{fig:diffmodels}, bottom row, we generated several samples from the prior of each of the current state-of-the-art generative models, conditioned on the caption ``A group of people walk on a beach with surf boards". While all of the samples look sharp, the images generated by LAPGAN look more noisy and it is harder to make out definite objects, whereas the images generated by variational models trained with least squares cost function have a watercolor effect on the images. We found that the quality of generated samples was similar among different variants of conditional DRAW models.

As for the quantitative comparison of different models, we first compare the performances of the model trained with variational methods. We rank the images in the test set conditioned on the captions based on the variational lower bound of the log-probabilities 
and then report the Precision-Recall metric as an evaluation of the quality of the generative model (see Table~\ref{tab:results}.). Perhaps unsurprisingly, generative models did not perform well on the image retrieval task. To deal with the large computational complexity involved in looping through each test image, we create a shortlist of one hundred images including the correct one, based on the images having the closest 
Euclidean distance in the last fully-connected feature space of a VGG-like model~\citep{simonyan_convnet} trained on the CIFAR dataset\footnote{The architecture of the model is described here \url{http://torch.ch/blog/2015/07/30/cifar.html}. The shortlist of test images used for evaluation can be downloaded from \url{http://www.cs.toronto.edu/~emansim/cap2im/test-nns.pkl}.} \citep{krizhevsky_cifar}. 
Since there are ``easy'' images for which the model assigns high log-probabilities independent of the query caption, 
we instead look at the ratio of 
the likelihood of the image conditioned on the sentence to the likelihood of the image conditioned on the mean sentence representation in the training set, following the retrieval protocol of~\citep{kiros_captions}.
We also found that the lower bound on the test log-probabilities decreased for sharpened images, and that sharpening considerably hurt the retrieval results. Since sharpening changes the statistics of images, the estimated log-probabilities of image pixels is not necessarily a good metric.
Some examples of generated images before and after sharpening are shown in Appendix~C.

Instead of calculating error per pixel, we turn to a smarter metric, the Structural Similarity Index (SSI) \citep{wang_ssi}, which incorporates luminance and contrast masking into the error calculation. Strong inter-dependencies of closer pixels are also taken into account and the metric is calculated on small windows of the images. Due to independence property of test captions, we sampled fifty images from the prior of each generative model for every caption in the test set in order to calculate SSI. As you can see on Table~\ref{tab:results}, SSI scores achieved by variational models were higher compared to SSI score achieved by LAPGAN.

\begin{table}[!t]
\vspace{-0.3in}
\begin{center}
\begin{tabulary}{\linewidth}{c || c c c c c || c c}
\hline
\multicolumn{7}{c}{\textbf{Microsoft COCO (prior to sharpening)}} \\
\hline
& \multicolumn{5}{c||}{Image Retrieval} & \multicolumn{1}{c}{Image Similarity} \\
\textbf{Model} & \textbf{R@1} & \textbf{R@5} & \textbf{R@10} & \textbf{R@50} & \textbf{Med r} & \textbf{SSI}\\
\hline
\hline
LAPGAN & - & - & - & - & - & 0.08 $\pm$ 0.07 \\ 
\hline
Fully-Conn VAE & 1.0 & 6.6 & 12.0 & 53.4 & 47 & 0.156 $\pm$ 0.10 \\
Conv-Deconv VAE & 1.0 & 6.5 & 12.0 & 52.9 & 48 & 0.164 $\pm$ 0.10 \\ 
skipthoughtDRAW & 2.0 & 11.2 & 18.9 & 63.3 & 36 & 0.157 $\pm$ 0.11 \\
noalignDRAW & 2.8 & 14.1 & 23.1 & 68.0 & 31 & 0.155 $\pm$ 0.11 \\
alignDRAW & 3.0 & 14.0 & 22.9 & 68.5 & 31 & 0.156 $\pm$ 0.11 \\
\end{tabulary}
\end{center}
\caption{Retrieval results of different models. 
\textbf{R@K} is Recall@K
    (higher is better). \textbf{Med} {\it r} is the median rank (lower is better). 
    \textbf{SSI} is Structural Similarity Index, which is between $-1$ and $1$ 
    (higher is better).
}
\label{tab:results}
\vspace{-0.1in}
\end{table}

\section{Discussion}
\vspace{-0.1in}
In this paper, we demonstrated that the alignDRAW model, a combination of a recurrent variational autoencoder with an alignment model over words, succeeded in generating images that correspond to a given input caption. By extensively using attentional mechanisms, our model gained several advantages. Namely, the use of the visual attention mechanism allowed us to decompose the problem of image generation into a series of steps instead of a single forward pass, while the attention over words provided us an insight whenever our model failed to generate a relevant image. Additionally, our model generated images corresponding to captions which generalized beyond the training set, such as sentences describing novel scenarios which are highly unlikely to occur in real life.

Because the alignDRAW model tends to output slightly blurry samples, we augmented the model with a sharpening post-processing step in which GAN generated edges which were added to the alignDRAW samples. Unfortunately, this is not an ideal solution due to the fact that the whole model was not trained in an end-to-end fashion. Therefore a direction of future work would be to find methods that can bypass the separate post-processing step and output sharp images directly in an end-to-end manner. 

{\small
{\bf Acknowledgments:}
This work was supported by Samsung and IARPA, Raytheon BBN Contract No. D11PC20071.
We would like to thank developers of Theano \citep{theano}, the authors of \citep{denton_lapgan} for open sourcing their code, and Ryan Kiros and Nitish Srivastava for helpful discussions. 
}

{\small
\bibliography{main}
\bibliographystyle{iclr-style/iclr2016_conference}
}

\newpage
\appendix
\pdfoutput=1

\section*{Appendix A: MNIST With Captions}

\begin{figure}[!t]
\captionsetup[subfigure]{labelformat=empty}
\begin{center}
\includegraphics[width=0.99\textwidth]{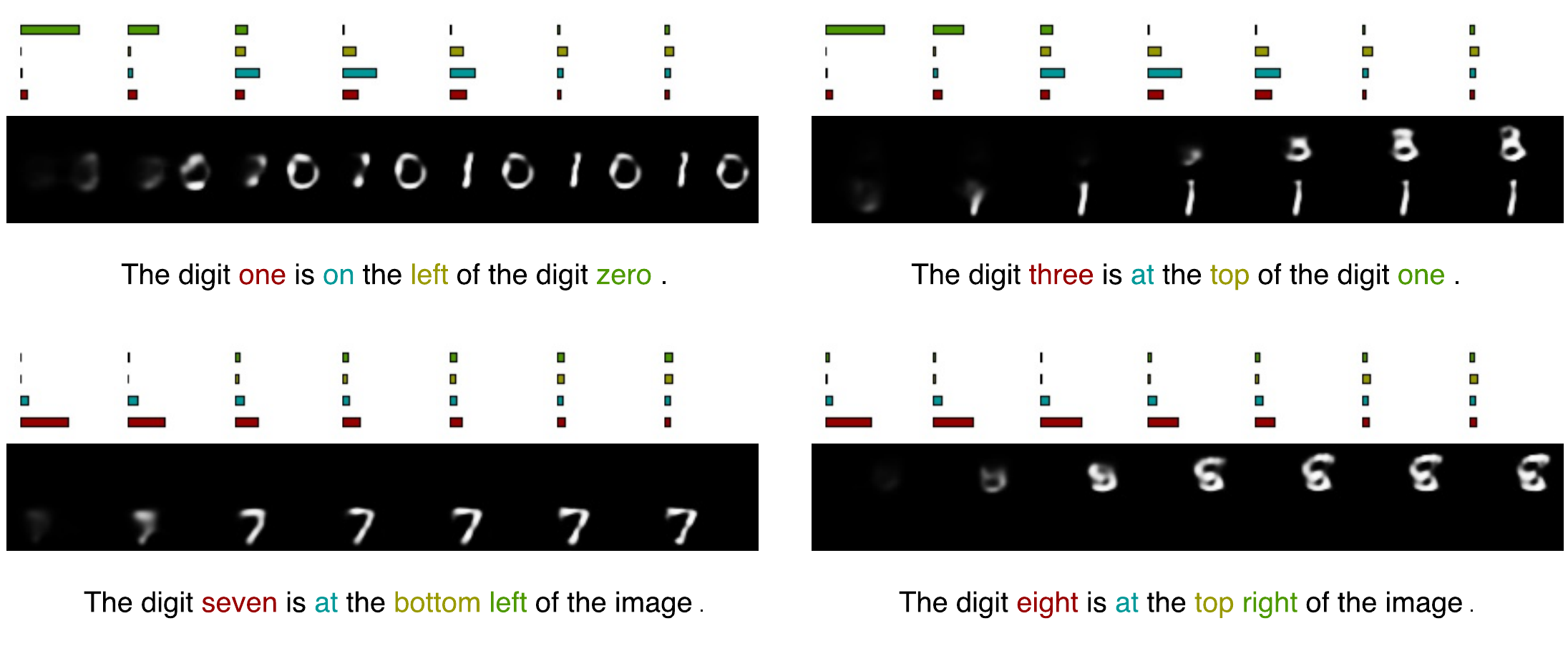}\quad
\end{center}
\caption{Examples of generating $60 \times 60$ MNIST images corresponding to respective captions. The captions on the \textbf{left column} were part of the training set. The digits described in the captions on the \textbf{right column} were hidden during training for the respective configurations.}
\label{fig:figmnist}
\vspace{-0.3cm}
\end{figure}

As an additional experiment, we trained our model on the MNIST dataset with artificial captions. Either one or two digits from the MNIST training dataset were placed on a $60 \times 60$ blank image. One digit was placed in one of the four (top-left, top-right, bottom-left or bottom-right) corners of the image. Two digits were either placed horizontally or vertically in non-overlapping fashion. The corresponding artificial captions specified the 
identity of each digit along with their relative positions, e.g. ``The digit three is at the top  
of the digit one'', or ``The digit seven is at the bottom left of the image''.  

The generated images together with the attention alignments are displayed in Figure~\ref{fig:figmnist}. The model correctly displayed the specified digits at the described positions and even managed to generalize 
reasonably well to the configurations that were never present during training.
In the case of generating two digits, the model would dynamically attend 
to the digit in the caption it was drawing at that particular time-step. 
Similarly, in the setting where the caption specified only a single digit, the model would correctly 
attend to the digit in the caption during the whole generation process. 
In both cases, the model placed small attention values on the words describing the position of digits in the images.

\section*{Appendix B: Training Details}
\label{sec:training_details}

\subsection*{Hyperparameters}

Each parameter in alignDRAW was initialized by sampling from a Gaussian distribution with 
mean $0$ and standard deviation $0.01$. 
The model was trained using \textit{RMSprop} with an initial learning rate of $0.001$. 
For the Microsoft COCO task, we trained our model for $18$ epochs. The learning rate was 
reduced to $0.0001$ after $11$ epochs.
For the MNIST with Captions task, the model was trained for 150 epochs and the 
learning rate was reduced to $0.0001$ after 110 epochs. During each epoch, 
randomly created $10,000$ training samples were used for learning. 
The norm of the gradients was clipped at $10$ during training 
to avoid the exploding gradients problem. 

We used a vocabulary size 
of $K = 25323$ and $K = 22$ for the Microsoft COCO and MNIST with Captions datasets respectively. 
All capital letters in the words were converted to small letters as a preprocessing step. 
For all tasks, the hidden states $\overrightarrow{h}^{lang}_{i}$ and $\overleftarrow{h}^{lang}_{i}$ 
in the language model had $128$ units. Hence the dimensionality of the 
concatenated state of the Bidirectional LSTM
$\hlang_{i} = [\overrightarrow{h}^{lang}_{i}, \overleftarrow{h}^{lang}_{i}]$ was~256. 
The parameters in the \textit{align} operator (\Eqref{eq:align2}) 
had a dimensionality of $l = 512$, so that 
$\vv \in \mathbb{R}^{512}$, $U \in \mathbb{R}^{512 \times 256}$, $W \in \mathbb{R}^{512 \times n^{gen}}$ and $b \in \mathbb{R}^{512}$. 
The architectural configurations of the alignDRAW models are shown in Table~\ref{tab:drawhyper}.

\begin{table}[!t]
\begin{center}
\begin{tabulary}{\linewidth}{c | c c c c c c}
\hline
 & & \multicolumn{3}{c}{\textbf{alignDRAW Model}}  & &  \\
\hline
Task & \#glimpses & Inference  & Generative  & Latent & Read Size & Write Size\\
     & \ T & Dim of $h^{infer}$ & Dim of $h^{gen}$  & Dim of $Z$  & $p$ & $p$\\
\hline
MS COCO & 32 & 550 & 550 & 275 & 9 & 9\\
MNIST   & 32 & 300 & 300 & 150 & 8 & 8\\
\end{tabulary}
\caption{The architectural configurations of alignDRAW models.}
\label{tab:drawhyper}
\end{center}
\end{table}

The GAN model used for sharpening had the same configuration as the $28 \times 28$ model trained by \cite{denton_lapgan} on the edge residuals of the CIFAR dataset. The configuration can be found at \url{https://gist.github.com/soumith/e3f722173ea16c1ea0d9}. The model was trained for $6$ epochs.

\subsection*{Evaluation}
Table~\ref{tab:nll} shows the estimated variational lower bounds on the average train/validation/test 
log-probabilities.
Note that the alignDRAW model does not suffer much from overfitting. 
The results substantially worsen after sharpening test images.

\begin{table}[!h]
\begin{center}
\begin{tabulary}{\linewidth}{c | c c c c}
\hline
Model & Train & Validation  & Test  & Test  (after sharpening)\\
\hline
skipthoughtDRAW & -1794.29 & -1797.41 & -1791.37 & -2045.84 \\
noalignDRAW & -1792.14 & -1796.94 & -1791.15 & -2051.07 \\
alignDRAW & -1792.15 & -1797.24 & -1791.53 & -2042.31
\end{tabulary}
\caption{The lower bound on the average test log-probabilities of conditional DRAW models, trained on the Microsoft COCO dataset.}
\label{tab:nll}
\end{center}
\end{table}

\newpage
\section*{Appendix C: Effect of Sharpening Images.}
\label{sec:post_processing}
\vspace{-0.1in}
Some examples of generated images before (top row) and after (bottom row) sharpening images 
using an adversarial network trained on residuals of a Laplacian 
pyramid conditioned on the skipthought vectors of the captions.
\begin{figure}[!h]
\captionsetup[subfigure]{labelformat=empty}
\vspace{-0.15in}
\begin{center}
\subfloat[]{\includegraphics[width=0.21\textwidth]{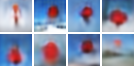}}\quad
\subfloat[]{\includegraphics[width=0.21\textwidth]{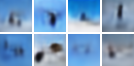}}\quad
\subfloat[]{\includegraphics[width=0.21\textwidth]{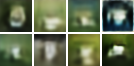}}\quad
\subfloat[]{\includegraphics[width=0.21\textwidth]{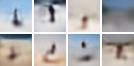}}\\
\vspace{-0.45cm}
\subfloat[
]{\includegraphics[width=0.21\textwidth]{figures/a-stop-sign-is-flying-in-blue-skies-sharp.png}}\quad
\subfloat[
]{\includegraphics[width=0.21\textwidth]{figures/a-herd-of-elephants-flying-in-the-blue-skies-sharp.png}}\quad
\subfloat[
]{\includegraphics[width=0.21\textwidth]{figures/a-toilet-seat-sits-open-in-the-grass-field-sharp.png}}\quad
\subfloat[
]{\includegraphics[width=0.21\textwidth]{figures/a-person-skiing-on-sand-clad-vast-desert-sharp.png}}\\
\vspace{-0.25cm}
\subfloat[]{\includegraphics[width=0.21\textwidth]{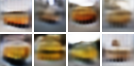}}\quad
\subfloat[]{\includegraphics[width=0.21\textwidth]{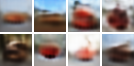}}\quad
\subfloat[]{\includegraphics[width=0.21\textwidth]{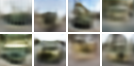}}\quad
\subfloat[]{\includegraphics[width=0.21\textwidth]{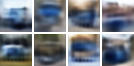}}\\
\vspace{-0.45cm}
\subfloat[
]{\includegraphics[width=0.21\textwidth]{figures/a-yellow-school-bus-parked-in-a-parking-lot-sharp.png}}\quad
\subfloat[
]{\includegraphics[width=0.21\textwidth]{figures/a-red-school-bus-parked-in-a-parking-lot-sharp.png}}\quad
\subfloat[
]{\includegraphics[width=0.21\textwidth]{figures/a-green-school-bus-parked-in-a-parking-lot-sharp.png}}\quad
\subfloat[
]{\includegraphics[width=0.21\textwidth]{figures/a-blue-school-bus-parked-in-a-parking-lot-sharp.png}}\\
\vspace{-0.25cm}
\subfloat[]{\includegraphics[width=0.21\textwidth]{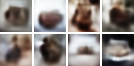}}\quad
\subfloat[]{\includegraphics[width=0.21\textwidth]{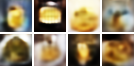}}\quad
\subfloat[]{\includegraphics[width=0.21\textwidth]{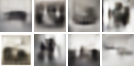}}\quad
\subfloat[]{\includegraphics[width=0.21\textwidth]{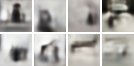}}\\
\vspace{-0.45cm}
\subfloat[
]{\includegraphics[width=0.21\textwidth]{figures/the-decadent-chocolate-dessert-is-on-the-table-sharp.png}}\quad
\subfloat[
]{\includegraphics[width=0.21\textwidth]{figures/a-bowl-of-bananas-is-on-the-table-sharp.png}}\quad
\subfloat[
]{\includegraphics[width=0.21\textwidth]{figures/a-vintage-photo-of-a-cat-sharp.png}}\quad
\subfloat[
]{\includegraphics[width=0.21\textwidth]{figures/a-vintage-photo-of-a-dog-sharp.png}}\\
\vspace{-0.25cm}
\subfloat[]{\includegraphics[width=0.21\textwidth]{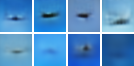}}\quad
\subfloat[]{\includegraphics[width=0.21\textwidth]{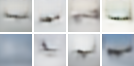}}\quad
\subfloat[]{\includegraphics[width=0.21\textwidth]{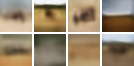}}\quad
\subfloat[]{\includegraphics[width=0.21\textwidth]{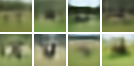}}\\
\vspace{-0.45cm}
\subfloat[
]{\includegraphics[width=0.21\textwidth]{figures/a-very-large-commercial-plane-flying-in-blue-skies-sharp.png}}\quad
\subfloat[
]{\includegraphics[width=0.21\textwidth]{figures/a-very-large-commercial-plane-flying-in-rainy-skies-sharp.png}}\quad
\subfloat[
]{\includegraphics[width=0.21\textwidth]{figures/a-herd-of-elephants-walking-across-a-dry-grass-field-sharp.png}}\quad
\subfloat[
]{\includegraphics[width=0.21\textwidth]{figures/a-herd-of-elephants-walking-across-a-green-grass-field-sharp.png}}\\
\vspace{-0.25cm}
\subfloat[]{\includegraphics[width=0.21\textwidth]{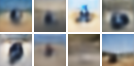}}\quad
\subfloat[]{\includegraphics[width=0.21\textwidth]{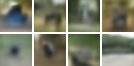}}\quad
\subfloat[]{\includegraphics[width=0.21\textwidth]{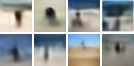}}\quad
\subfloat[]{\includegraphics[width=0.21\textwidth]{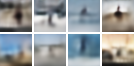}}\\
\vspace{-0.45cm}
\subfloat[
]{\includegraphics[width=0.21\textwidth]{figures/a-rider-on-a-blue-motorcycle-in-the-desert-sharp.png}}\quad
\subfloat[
]{\includegraphics[width=0.21\textwidth]{figures/a-rider-on-a-blue-motorcycle-in-the-forest-sharp.png}}\quad
\subfloat[
]{\includegraphics[width=0.21\textwidth]{figures/a-surfer-,-a-woman-,-and-a-child-walk-on-the-beach-sharp.png}}\quad
\subfloat[
]{\includegraphics[width=0.21\textwidth]{figures/a-surfer-,-a-woman-,-and-a-child-walk-on-the-sun-sharp.png}}\\
\end{center}
\caption{\small Effect of sharpening images.}
\end{figure}

\end{document}